\documentclass[11pt,a4paper]{article}
\usepackage[hyperref]{emnlp2018}
\usepackage{times}
\usepackage{latexsym}
\usepackage{graphicx}
\usepackage{CJK}
\usepackage{url}
\usepackage{makecell}
\usepackage{amsmath}
\usepackage{algorithm}
\usepackage{algorithmicx}
\usepackage{algpseudocode}
\usepackage{amsfonts,amssymb}
\usepackage{color,soul}
\usepackage{tikz}
\usepackage{environ}
\usepackage{xcolor}

\aclfinalcopy 
\def\aclpaperid{907} 

\newcommand*{\affmark}[1][*]{\textsuperscript{#1}}
\newcommand*{\email}[1]{\texttt{#1}}
\title{A Skeleton-Based Model for Promoting Coherence Among Sentences \\in Narrative Story Generation }

\author{
Jingjing Xu\affmark[1], \  
Xuancheng Ren\affmark[1], \ 
Yi Zhang\affmark[1], \
Qi Zeng\affmark[1], \
Xiaoyan Cai\affmark[2], \
Xu Sun\affmark[1]
\\ 
\affmark[1]MOE Key Lab of Computational Linguistics, School of EECS, Peking University  \\
\affmark[2]School of Automation, Northwestern Polytechnical University \\
\email{\{jingjingxu,renxc,zhangyi16,pkuzengqi,xusun\}@pku.edu.cn }\\
\email{xiaoyanc@nwpu.edu.cn}}

\date{}

\begin{document}
\maketitle
\begin{CJK}{UTF8}{song}

\begin{abstract}
Narrative story generation is a challenging problem because it demands the generated sentences with tight semantic connections, which has not been well studied by most existing generative models. To address this problem, we propose a  skeleton-based model to promote the coherence of generated stories. Different from traditional models that generate a complete sentence at a stroke, the proposed model first generates the most critical phrases, called skeleton, and then expands the skeleton to a complete and fluent sentence. The skeleton is not manually defined, but learned by a reinforcement learning method. Compared to the state-of-the-art models, our skeleton-based model  can generate significantly more coherent text according to human evaluation and automatic evaluation. The G-score is improved by 20.1\% in human evaluation.\footnote{The code is available at \url{https://github.com/lancopku/Skeleton-Based-Generation-Model}} 
\end{abstract}

\section{Introduction}

We focus on the problem of narrative story generation, a special kind of story generation~\cite{DBLP:conf/aaai/LiLJR13}. It requires systems to generate a narrative story based on a short description of a scene or an event, as shown in Table~\ref{example}. In general, a narrative story is described with several inter-related scenes. Different from traditional text generation tasks, this task is more challenging because it demands the generated sentences with tight semantic connections. Currently, most state-of-the-art approaches~\cite{DBLP:journals/corr/JainAMSLS17,DBLP:journals/corr/abs-1711-09724, faceboook18,DBLP:journals/corr/abs-1805-04869,unpaired-sentiment-translation} are largely based on Sequence-to-Sequence (Seq2Seq) models~\cite{SutskeverVL14_seq2seq}, which generate a sentence at a stroke in a left-to-right manner. 

\begin{figure*}[t] 
\centering
\includegraphics[width = 0.8\linewidth]{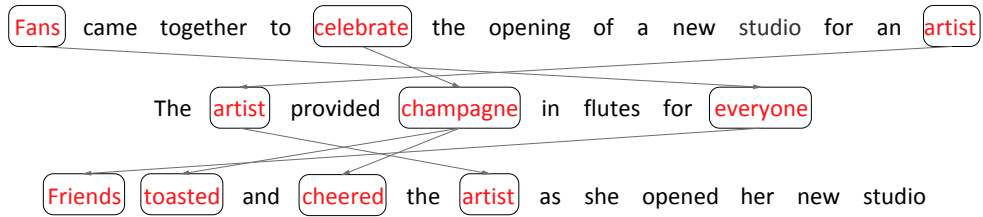}
\caption{The semantic dependency among sentences in a narrative story. It can be seen that the connection among sentences is mainly reflected through key phrases (shown in red). In this work, we regard such key phrases as a skeleton.  }
\label{samplecase} 
\vspace{-1\baselineskip}
\end{figure*}

However, we find it hard for these approaches to model the semantic dependency among sentences, which causes low-quality generated stories where the scenes are irrelevant. In fact, as shown in Figure~\ref{samplecase}, we observe that the connection among sentences is mainly reflected through key phrases, such as predicates, subjects, objects and so on. In this work, we regard the phrases that express the key meanings of a sentence as a skeleton. The other words (e.g., modifiers) not only are redundant for understanding semantic dependency, but also make the dependency sparse. Therefore, generating all information at a stroke makes it difficult  to learn the dependency of sentences. In contrast, the sentences written by humans are closely tied and the whole story is more coherent and fluent. It is mainly attributed to the way of human writing where we often first come up with a skeleton and then reorganize them into a fluent sentence.

\begin{table}[!t]
\footnotesize
\centering
    \begin{tabular}{p{0.9\linewidth}}
    \Xhline{1.2pt}
     
    \multicolumn{1}{c}{\textbf{Task Description}}\\ \hline
    Input: A short description of a scene or an event.\\
    Output: A relevant narrative story following the input. \\ \Xhline{1.2pt}

     \multicolumn{1}{c}{\textbf{Examples}}\\ \hline 
     
     \textbf{Input}: \textsl{Fans came together to celebrate the opening of a new studio for an artist.}\\
     \textbf{Output}: \textsl{The artist provided champagne in flutes for everyone. Friends toasted and cheered the artist as she opened her new studio.}\\\hline
    
     \textbf{Input}: \textsl{Last week I attended a wedding for the first time.}  \\
     \textbf{Output}: \textsl{There were a lot of families there. They were all taking pictures together. Everyone was very happy. The bride and groom got to ride in a limo that they rented.} \\\Xhline{1.2pt}
      

    \end{tabular}
    \caption{An illustration of narrative story generation. }
    \label{example}
\vspace{-1.5\baselineskip}
\end{table}

Therefore, motivated by the way of human writing, we propose a skeleton-based model to improve the coherence of generated text. The key idea is to first generate a skeleton and then expand the skeleton to a complete sentence.  As a simplified sentence representation, the skeleton can help machines learn the dependency of sentences by avoiding the interference of irrelevant information. Our model contains two parts: a skeleton-based generative module and a skeleton extraction module. 

The generative module consists of an input-to-skeleton component and a skeleton-to-sentence component. The input-to-skeleton component learns to associate inputs and skeletons, and the skeleton-to-sentence component learns to expand a skeleton to a sentence. In our model, a good skeleton that can capture key semantic information is a critical supervisory signal.
 
 The skeleton extraction module is used to generate sentence skeletons.
In real-world datasets, the human-annotated skeleton is usually unavailable. In addition, it is difficult to define the unified rules of extracting skeletons, because different sentences have different focuses. To address this problem, we build a skeleton extraction module to automatically explore sentence skeletons.  Considering the discrete choice of skeleton words causes the loss function to be non-differentiable, we use a reinforcement learning method to build the connection between the skeleton extraction module and the generative module. 

Our contributions are listed as follows:
\begin{itemize}
\item A skeleton-based model is proposed to promote the coherence of generated stories. 
\item The proposed model contains a skeleton-based generative module and a skeleton extraction module. Two modules are connected by a reinforcement learning method to automatically explore sentence skeletons.
\item The experimental results on automatic evaluation and human evaluation show that our model can generate significantly more coherent text compared to the state-of-the-art models.
\end{itemize}

\section{Related Work}

Strictly speaking, the story generation task requires systems to generate a story from scratch without any external materials. However, for simplification, many existing story generation models
rely on their given materials, such as short text descriptions~\cite{harrison2017toward,DBLP:journals/corr/JainAMSLS17}, visual images~\cite{charles2001character,huang2016visual}, and so on. Different from these studies, we get rid of external materials and consider the complete story generation task~\cite{DBLP:conf/acl/McIntyreL09}. For this task, the widely used models are based on Seq2Seq models. However, although they can generate a fluent sentence~\cite{jingjingxuemnlp18-01}, these models still perform badly on generating inter-related sentences, which are necessary for a coherent story. 


To address this problem, there are several models that build the mid-level sentence semantic representation to simplify the dependency among sentences.  \citet{entity} extract the entities in sentences, and combine the entity context and text context together when generating a target sentence. \citet{DBLP:conf/aaai/CaoWLL18} encode the words with specific pre-defined dependency labels to a mid-level sentence representation. \citet{DBLP:conf/aaai/MartinAWHSHR18} use additional knowledge bases to get a generalized sentence representation. \citet{DBLP:journals/corr/abs-1805-04871} use the bag-of-words which occur in all references as a representation of the correct translation. ~\citet{jingjingxuemnlp18-02} propose to use two auto-encoders to learn the semantic representation of utterance in dialogue. However, although these models reduce the dependency sparsity to some extent, the unified rules are non-flexible and tend to generate over-simplified representations, resulting in the loss of key information. 

Different from these models, we use a reinforcement learning method to automatically extract sentence skeletons for simplifying the dependency of sentences, rather than manual rules. Therefore, our proposed skeleton-based model is more flexible and can adaptively determine the appropriate granularity of sentence representations for a balance between keeping key semantics and simplifying sentence representations. 

\begin{figure}[t] 
\centering
\includegraphics[width = 0.8\linewidth]{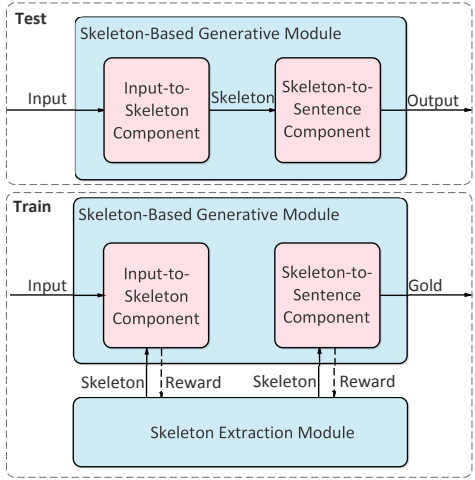}
\caption{An illustration of the proposed model. Top: Testing. Bottom: Training. The ``input'' means the existing text, including the source input and the already generated text. The ``skeleton'' means the skeleton of the output. The skeleton extraction module first extracts skeletons from gold outputs. The pairs of input-skeleton and skeleton-gold are used to train the input-to-skeleton component and the skeleton-to-sentence component. In return, the generative module can be used to evaluate the quality of extracted skeletons. Therefore, we use the feedback of the generative module to reward extracted skeletons. By cooperation, the two modules can promote each other until convergence. }
\label{structure} 
\vspace{-1.0\baselineskip}
\end{figure} 


\section{Skeleton-Based Model}
An overview of our proposed skeleton-based model is presented in Section~\ref{overview}. The details of the skeleton-based generative module and the skeleton extraction module are shown in Section~\ref{generative} and Section~\ref{sts}. The reinforcement learning method is explained in Section~\ref{rl}.

\subsection{Overview}\label{overview}
As shown in Figure~\ref{structure}, our model consists of two parts, a skeleton-based generative module $G_{\phi}$ and a skeleton extraction module $E_{\gamma}$. The generative module consists of an input-to-skeleton component and a skeleton-to-sentence component.

The generative module generates a story sentence by sentence. When decoding a sentence, the input-to-skeleton component first generates a skeleton based on the existing text, including the source input and the already generated text, and then the skeleton-to-sentence component expands and reorganizes the skeleton into a complete sentence. We keep running this process until the generative module generates an ending symbol.

In the training process, we first use a weakly supervised method to assign the skeleton extraction module with initial extraction ability. Then, we use extracted skeletons to train the input-to-skeleton component and the skeleton-to-sentence component. In return, the generative module can be used to evaluate the quality of extracted skeletons. Therefore, we use the feedback of the generative module to reward extracted skeletons. The reward refines the skeleton extraction module. The improved skeleton extraction module further enhances the generative module. By cooperation, the two modules can promote each other until convergence.

\subsection{Skeleton-Based Generative Module}\label{generative}
The skeleton-based generative module $G_{\phi}$ consists of two parts: an input-to-skeleton component $Q_{\alpha}$ and a skeleton-to-sentence component $D_{\theta}$.

\subsubsection{Input-to-Skeleton Component}
The input-to-skeleton component $Q_{\alpha}$ builds on a Seq2Seq structure with a hierarchical encoder~\cite{LiLJ15_hierencoder} and an attention-based decoder~\cite{BahdanauCB14_attention}. It is responsible for learning the dependency between inputs and skeletons. In the encoding process, we first obtain sentence representations via a word-level Long Short Term Memory (LSTM) network~\cite{hochreiter1997long}, and then generate a compressed vector $h$ via a sentence-level LSTM network. Finally, given the compressed vector $h$, the attention-based decoder is responsible for imagining a skeleton. 

Given the training pair of input $\boldsymbol{c}$ and skeleton $\boldsymbol{s}=\{s_1,\cdots,s_i,\cdots,s_T\}$, the cross-entropy loss is computed as
\begin{equation}\label{em_0}
L_{\alpha} = -\sum_{i=1}^{T}P_{Q}(s_i|\boldsymbol{c}, \alpha) 
\end{equation}
where $\alpha$ refers to the parameters of the input-to-skeleton component. The skeleton $\boldsymbol{s}$ is extracted by the skeleton extraction module. The extracting details will be introduced in Section~\ref{sts}.

\subsubsection{Skeleton-to-Sentence Component}
The skeleton-to-sentence component $D_{\theta}$ builds on a Seq2Seq structure. Both the encoder and the decoder are one-layer LSTM networks with the attention mechanism. Given a skeleton $\boldsymbol{s}$, the encoder first generates a compressed representation, which is then used to generate a detailed and polished sentence via the decoder.

Given the training pair of skeleton $\boldsymbol{s}$ and target sentence  $\boldsymbol{y} = \{y_1,\cdots,y_i,\cdots,y_M\}$, the cross-entropy loss is computed as
\begin{equation}\label{em_1}
L_{\theta} = -\sum_{i=1}^{M}P_{D}(y_i|\boldsymbol{s}, \theta) 
\end{equation}
where $\theta$ refers to the parameters of the skeleton-to-sentence component.

\subsection{Skeleton Extraction Module}\label{sts}

Given a sentence $\boldsymbol{x}$, the skeleton extraction module $E_{\gamma}$  is responsible for extracting its skeleton that only preserves the key information. Specially, we use the Seq2Seq model with the attention mechanism as the implementation. Both the encoder and the decoder are based on LSTM structures.

Since the extracted skeletons are treated as supervisory signals for the generative module, the extraction ability needs to be initialized. 
To pre-train the skeleton extraction module, we propose a weakly supervised method. We reformulate skeleton extraction as a sentence compression problem and use a sentence compression dataset to train this module. In sentence compression, the compressed sentence is required to be grammatical and convey the most important information. From the aspect of keeping important information, the sentence compression dataset can be used to help the training of the skeleton extraction module. However, since the style of the sentence compression dataset is very different from that  of the narrative story dataset, it is difficult for the pre-trained module to give narrative text accurate compression results. Therefore, the supervisory signals are noisy and need to be further improved.

Given the training pair of the original text $\boldsymbol{x}$ and the compressed version $\boldsymbol{s}=\{s_1, \cdots, s_i, \cdots, s_T\}$, we use the following cross-entropy loss to pre-train $E_{\gamma}$:
\begin{equation}\label{em}
L_{\gamma} = -\sum_{i=1}^{T}P_{E}(s_i|\boldsymbol{x}, \gamma) 
\end{equation}
where $\gamma$ is the parameters of the skeleton extraction module.

\begin{algorithm}[t]

\caption{The reinforcement learning method for training the generative module $G_{\phi}$ and the skeleton extraction module $E_{\gamma}$.}
 
\small
  \begin{algorithmic}[1]

      \State Initialize the generative module $G_{\phi}$ and the skeleton extraction module $E_{\gamma}$ with random weights $\phi$, $\gamma$ 
      \State Pre-train $E_{\gamma}$ using MLE based on Eq. \ref{em} 
    \For{each iteration $j=1,2,..., J$}  
        \State Generate a skeleton $\boldsymbol{s}_{j}$ based on $E_{\gamma}$ 
        \State Given $\boldsymbol{s}_{j}$, train $G_{\phi}$ based on Eq. \ref{em_0} and Eq. \ref{em_1}.
        \State Compute the reward $R_c$  based on Eq.~\ref{combined_reward} 
        \State Compute the gradient of $E_{\gamma}$ based on Eq. \ref{learning}
        \State Update the model parameter $\gamma$
    \EndFor  
     
  \end{algorithmic} 
  \label{code}
\end{algorithm}  

\subsection{Reinforcement Learning Method}\label{rl}

We propose a reinforcement learning method to build the connection between the skeleton extraction module and the skeleton-based generative module for exploring better skeletons. 
The detailed training process is shown in Algorithm~\ref{code}.



Due to the discrete choice of words in skeletons, the loss is no longer differentiable over the skeleton extraction module. Therefore, we use policy gradient ~\cite{SuttonMSM99_policygrad} to train the skeleton extraction module.

First, we calculate a reward $R_{c}$ based on the feedback of the generative module. The details of calculation process will be introduced in Section~\ref{reward-sec}. Then, we optimize the parameters through policy gradient by maximizing the expected reward to train the skeleton extraction module. According to the policy gradient theorem, the gradient for the skeleton extraction module is
\begin{equation}\label{learning}
\begin{split}
\nabla J(\gamma)= \mathbb{E}[R_{c} \cdot \nabla\log(P_{E}(\boldsymbol{s} |\boldsymbol{x}),\gamma)]
\end{split}
\end{equation}
where $\boldsymbol{x}$ is the original sentence, $\boldsymbol{s} $ is the skeleton generated by a sampling mechanism.

\subsubsection{Reward}\label{reward-sec}

To design an appropriate rewarding function, there is a critical question that needs to be considered: what will good/bad skeletons bring to the generative module. 

We first define what is a good (or bad) skeleton.
A good skeleton is expected to contain all key information and ignore other  information.
In contrast, the skeletons that contain too much detailed information or lack necessary information are considered as bad skeletons and should be punished.
For ease of analysis, we classify possible scenarios into three categories:
good skeletons, incomplete skeletons, and redundant skeletons.

If a skeleton contains too little information,  it will get harder for the skeleton-to-sentence component to reconstruct the original sentence based on the skeleton. Therefore, the cross-entropy loss of this example will be higher compared with other skeleton-sentence pairs.

If a skeleton contains too much redundant information, the input-skeleton relation will be sparse. Therefore, the cross-entropy loss of this example will be higher compared with other input-skeleton pairs. 

For a good skeleton that contains an appropriate amount of information, it will benefit the two components and will get balanced losses from the two components. 

Therefore, to encourage good skeletons and punish bad skeletons, we use the multiplication of the cross-entropy loss in the input-to-skeleton component and that in the skeleton-to-sentence component as the reward:
\begin{equation}\label{combined_reward}
\begin{split}
R_{c} = [K-(R_{1}\times R_{2})^{\frac{1}{2}}]
\end{split}
\end{equation}
where $K$ is the upper bound of the reward, $R_{1}$ and $R_{2}$ are cross-entropy losses in the input-to-skeleton component and the skeleton-to-sentence component, respectively. Only if two components both output lower cross-entropy losses, the extracted skeleton can be rewarded. 

\section{Experiment}
In this section, we evaluate our model on a narrative story generation dataset. We first introduce the dataset, the training details, the baselines, and the evaluation metrics. Then, we compare our model with the state-of-the-art models. Finally, we show the experimental results and provide the detailed analysis.

 
\subsection{Dataset}

We use the recently introduced visual storytelling
dataset~\cite{huang2016visual} in our experiments. This dataset contains the pairs of photo sequences and the associated coherent narrative of events through time written by humans. We only use the text data for our experiments. In our version of narrative story generation, the model should generate a coherent story based on a given sentence. We build a new dataset for this task by splitting the data into two parts. In each story, we take the first sentence as the input text, and the following sentences as the target text. The processed dataset contains 40153, 4990, and 5054 stories for training, validation, and testing, respectively. The maximum number of sentences in each story is 6. In total, the number of training sentences is over 20K and the number of training words is over 2M. 
 
To pre-train the skeleton extraction module, we use a sentence compression dataset~\cite{DBLP:conf/emnlp/FilippovaA13}. In this dataset, every compression is a subsequence of tokens from the input. The dataset contains 16999, 1000, and 1998 pairs for training, validation, and testing, respectively.

\subsection{Baselines}

We compare our proposed model with the following the state-of-the-art models. 

\textbf{Entity-Enhanced Seq2Seq Model  (EE-Seq2Seq)}~\cite{entity}. It regards entities as important context needed for coherent stories. When decoding a sentence, it combines entity context and text context together to reduce dependency sparsity. 



\textbf{Dependency-Tree Enhanced Seq2Seq Model (DE-Seq2Seq)}~\cite{DBLP:conf/aaai/CaoWLL18}. It defines some manual rules based on dependency parsing labels to find a simplified sentence representation. Following this work, we treat the extracted words based on the predefined rules as the skeleton.

\textbf{Generalized-Template Enhanced Seq2Seq Model  (GE-Seq2Seq)}~\cite{DBLP:conf/aaai/MartinAWHSHR18}. It takes advantages of existing knowledge bases to get a generalized sentence representation. Following this work, we treat the generalized sentence representation as the skeleton.

\subsection{Training Details}

For narrative story generation, we set the number of generated sentences to 6 with the maximum length of 40 words for each generated sentence. Based on the performance on the validation set, we set the hidden size to 128, embedding size to 50, vocabulary size to 20K, and batch size to 10 for the proposed model and the state-of-the-art models. We use the Adagrad~\citep{DuchiHS11_adagrad} optimizer with the initial learning rate 0.6. All of the gradients are clipped when the norm exceeds 2. Both the generative module and skeleton extractive module are pre-trained for 30 and 40 epochs before reinforcement learning. The $K$ in Equation~\ref{combined_reward} is set to 1. Due to the lack of annotated entities and dependency parsing labels, we use a popular natural language processing toolkit, Spacy\footnote{https://spacy.io/}, to extract entities and dependency parsing labels in the EE-Seq2Seq and DE-Seq2Seq models.

\subsection{Evaluation Metrics}
We conduct two kinds of evaluations in this work, automatic evaluation and human evaluation. The details of evaluation metrics are shown as follows.

\subsubsection{Automatic Evaluation}

Following the previous work~\cite{DBLP:conf/emnlp/LiMRJGG16,DBLP:conf/aaai/MartinAWHSHR18}, we use the BLEU score to measure the quality of generated text.  BLEU~\cite{DBLP:conf/acl/PapineniRWZ02} is originally designed to automatically judge the machine translation quality. The key point is to compare the similarity between the results created by the machine and the references provided by the human. Currently, it is widely used in many generation tasks, such as dialogue generation, story generation, summarization, and so on. For precise results, we remove all stop words, like ``the'', ``a'', before computing BLEU scores.

\subsubsection{Human Evaluation}

Although the quantitative evaluation generally indicates the quality of generated stories, it can not accurately evaluate the generated text. 
Therefore, we also perform a human evaluation on the test set. We randomly choose 100 items for human evaluation. Each item contains the stories generated by different models given the same source sentence. The items are distributed to the annotators who have no knowledge about which model the story is from. It is important to note that all the annotators have linguistic background. They are asked to score the generated stories in terms of fluency and coherence. Fluency represents whether each sentence in the generated story is correct in grammar. Coherence evaluates whether the generated story is coherent. The score ranges from 1 to 10 (1 is very bad and 10 is very good). To evaluate the overall performance, we use the geometric mean of fluency and coherence as an evaluation metric.


\begin{table}[t]

\footnotesize
\centering
    \begin{tabular}{l|c}
    \hline
     Models &BLEU  \\ \hline
    EE-Seq2Seq & 0.0029 \\
    DE-Seq2Seq&0.0027\\
    GE-Seq2Seq & 0.0022\\\hline
    Proposed Model &\textbf{0.0042 (+44.8\%)} \\\hline
   
    \end{tabular}
    \caption{Automatic evaluations of the proposed model and the state-of-the-art models. }
    \label{tab:automatic}
    
\end{table}


\begin{table}[t]
\centering
\footnotesize
\setlength{\tabcolsep}{2.5pt}
    \begin{tabular}{l|c|c|c}
    \hline
     Models& Fluency & Coherence & G-Score \\ \hline
    EE-Seq2Seq& 6.28 & 5.14 & 5.68\\
    DE-Seq2Seq& 8.48 & 3.54 & 5.48\\
    GE-Seq2Seq& 9.48 & 3.58 &5.82\\\hline
    Proposed Model  & 8.69 & 5.62 &\textbf{6.99 (+20.1\%)}\\
    \hline
    \end{tabular}
    \caption{Human evaluations of the proposed model and the state-of-the-art models. }
    \label{tab:human}

\end{table}

\subsection{Experimental Results}

Table~\ref{tab:automatic} shows the results of automatic evaluation. The proposed model performs the best according to BLEU. In particular, the differences between the existing state-of-the-art models are within 0.07, while the proposed model supersedes the best of them by 0.13.


As we previously explained, the best evaluation for narrative story generation is human evaluation. The human evaluation results are listed in Table~\ref{tab:human}.\footnote{The inter-annotator agreement is satisfactory considering the difficulty in the human evaluation. The Pearson's correlation coefficient is 0.37 on coherence and 0.26 on fluency, with $p<0.0001$.} 

As for fluency, the proposed model receives the score of 8.69, second to the GE-Seq2Seq model. It is expected that the generalized templates can constrain the search space in generation and the model achieves higher fluency by loss of expressive power. In particular, we find that only 0.48\%, 1.01\%, and 1.20\% of the unigrams, bigrams, and trigrams are unique in the stories generated by the GE-Seq2Seq model, while the percentages are 3.16\%, 15.33\%, and 29.67\% in the stories generated by our proposed model. Nonetheless, the proposed model outperforms the other two existing models by a substantial margin. In terms of coherence, the proposed model is better than all the existing models. We need to point out that the GE-Seq2Seq model is scored the lowest in coherence, while highest in fluency. It indicates that the GE-Seq2Seq model does not learn the dependency among sentences effectively, which results from the constraint of the templates. It also needs to be noted that the models are all scored below 6 in coherence, meaning that there is still a long way to go before the generated stories satisfy the requirement of humans. Overall, the proposed model is arguably better than the existing models in that it achieves a balance between coherence and fluency, with a G-score improvement of 20.1\%.

Table~\ref{samplecases} presents the examples generated by different models. Compared with the existing models, the sentences generated by our proposed model are connected more logically. For the EE-Seq2Seq model, while it connects \textit{park} with \textit{plants} and \textit{rocks} successfully (4th ex.), it insists on telling \textit{getting married} when it sees \textit{[male]} or \textit{[female]} (1st and 2nd ex.). Such examples suggest that some entities (e.g. \textit{park}) embody semantics more independently, while for others (e.g. \textit{male}), we have to associate them in the specific context. The rest of the models try to generalize the target sentences. The DE-Seq2Seq model uses the core dependency arguments as the skeleton. However, the results demonstrate the generated sentences are quite irrelevant. The sentence may have links such as \textit{walked through} to \textit{came out} (1st ex.), but the objects in the generated stories are hardly related. The GE-Seq2Seq model replaces the specific words with more general concepts and generates some good examples, e.g. the second one in the table. However, there can be over-generalizations. For example, as for the third example, the GE-Seq2Seq model associates \textit{driving} with \textit{car show}, causing the incoherent description. In the last example, the generated story completely diverges from the input. These results prove the drawbacks of static rule-based skeletons. The proposed model uses a skeleton extraction module to adaptively determine the appropriate granularity of skeletons. The skeleton keeps the main semantic of a sentence, which can be a whole sentence, phrases, or even segments. It makes the model learn the dependency of sentences more effectively so that the generated stories are much more coherent.

\begin{table}[!t]
\footnotesize
\centering
    \begin{tabular}{p{0.95\linewidth}}
    \hline
       \textbf{Input 1}: \textsl{[Female] and her friend had a night out on the town.}\\
       \textbf{EE-Seq2Seq}:  \textsl{They were very happy to see us. [Male] and [female] were getting married today.}    \\  
      \textbf{DE-Seq2Seq}: \textsl{We walked through the streets of the city and found a place to sit. The whole family came out to support their families.}\\
     \textbf{GE-Seq2Seq}: \textsl{My friends and I went to a bar last night. [Female] was so happy to be there.}\\
    \textbf{Proposed Model}: \textsl{They went to the bar. They had a great time. }\\ 
    \hline


    \textbf{Input 2}: \textsl{[Male] and [male] are talented artists. }\\
       \textbf{EE-Seq2Seq}:  \textsl{They were very happy to see us. [Male] and [female] were getting married today. }    \\  
      \textbf{DE-Seq2Seq}:\textsl{The kids get to play with the kids. We had a great time.}\\
     \textbf{GE-Seq2Seq}:\textsl{The art was very subjective. The artist had a keen eye.}\\
    \textbf{Proposed Model}: \textsl{They are discussing the importance of the lecture. }\\ 
    \hline

     \textbf{Input 3}: \textsl{The bride was driving to the wedding.}\\
       \textbf{EE-Seq2Seq}:  \textsl{The family was so happy to be there. They were very happy to see us.}    \\  
      \textbf{DE-Seq2Seq}: \textsl{The view from the top was amazing. There were a lot of people there.}\\
     \textbf{GE-Seq2Seq}: \textsl{I went to the car show last weekend. People are getting ready to go. }\\
    \textbf{Proposed Model}: \textsl{The groom and groomsmen watched the bride. They kissed and shared their pictures.}\\ 
    \hline


    \textbf{Input 4}: \textsl{The park was filled with beauty.}\\
       \textbf{EE-Seq2Seq}:  \textsl{The plants were so beautiful. The rocks were very pretty. }    \\  
     \textbf{DE-Seq2Seq}: \textsl{The family was so happy to be there. }\\
     \textbf{GE-Seq2Seq}: \textsl{There were a lot of people there. We had a great time.}\\
     \textbf{Proposed Model}: \textsl{The trees were very sparse. There were also some flowers.}\\ 
    \hline


    \end{tabular}
    \caption{Examples generated by the proposed model and the state-of-the-art models. }
    \label{samplecases}

\end{table}


\begin{table}[t]
\footnotesize
\centering
    \begin{tabular}{l|c}
    \hline
    Models &BLEU  \\ \hline
    Seq2Seq &0.0028 \\
    +Skeleton Extraction Module&0.0029\\
    +Reinforcement Learning &\textbf{0.0042} \\\hline 
    \end{tabular}
    \caption{Automatic evaluations of key components.  }
    \label{tab:increment}
    
\end{table}


\begin{table}[t]
\centering
\footnotesize
\setlength{\tabcolsep}{1.5pt}
    \begin{tabular}{l|c|c|c}
    \hline
     Models& Fluency & Coherence & G-Score \\ \hline
    Seq2Seq& 7.54 & 4.98 & 6.13\\
   +Skeleton Extraction  Module& 7.26 & 4.32 & 5.60 \\
   +Reinforcement Learning& 8.69 & 5.62& \textbf{6.99}\\\hline
    \end{tabular}
    \caption{Human evaluations of the key components. }
    \label{tab:ab-human}
    
\end{table}

\subsection{Incremental Analysis}

In this section, we conduct a series of experiments to evaluate the contributions of our key components. The results are listed in Table~\ref{tab:increment}. The  Seq2Seq model is scored the lowest according to BLEU. With the skeleton extraction module, the BLEU score is slightly improved, which suggests that the model starts to learn the connection of longer segments. Finally, with reinforcement learning, the BLEU score significantly overpasses the Seq2Seq model by 40\%.

Table~\ref{tab:ab-human} shows the human evaluation results. The slight improvement with the skeleton extraction module in BLEU reflects as the decreases in both fluency and coherence. It suggests the necessity of human evaluation.
The decreased results can  be explained by the fact that the style of the dataset for pre-training the skeleton extraction module is very different from the narrative story dataset. While it may help extract some useful skeletons, it is likely that many of them are not suitable for learning the dependency of sentences. Finally, when the skeleton extraction module is trained on the target domain using reinforcement learning, the human evaluation is improved significantly by 14\% on G-score.

Table~\ref{tab:ex-sk} further shows the results of the skeleton extraction module. As we can see, the module keeps only the essential parts of the sentence. Most of the adjectival phrases and adverbial phrases are removed. Furthermore, we can find that for longer sentences that contain too detailed information, it only extracts the key information. For shorter sentences where all information is necessary, it choose to keep all words.  It proves that the skeleton extraction module is effective and is expert in only removing detailed information that is not needed.


Furthermore, it is not quite surprising to see that on our dataset, the Seq2Seq model beats the existing state-of-the-art models (DE-Seq2Seq and GE-Seq2Seq) in human evaluation and automatic evaluation. It is mainly attributed to the over-simplification of sentences. For narrative sentences, the key information is usually expressed in a complicated way. It can be a segment, a phrase, or a whole sentence. The simple rules lead to the excessive loss of key information while our proposed model can adaptively determine the appropriate granularity. 

\begin{table}[t]
\footnotesize
\centering
    \begin{tabular}{p{0.90\linewidth}}
    \hline
         1) \textsl{\textcolor{red}{There was a small power station} on the side of the building.}\\ 
         2) \textsl{\textcolor{red}{The lady} wearing the pink shirt \textcolor{red}{decided to stop playing the video} and chatted with other guests.} \\ 
         3) \textsl{At the end of the night, \textcolor{red}{guests taking pictures} before saying goodbye to each other.}\\
            4) \textsl{Afterwards, \textcolor{red}{we celebrated} with some drinks and watched the rest of the parade.}\\
            5) \textsl{A few miles away was a \textcolor{red}{lake} that \textcolor{red}{we} really \textcolor{red}{enjoyed watching.} }
            
            6) \textsl{Some of the guests \textcolor{red}{partied harder} than others.}\\
            7) \textsl{\textcolor{red}{The bride was driving to the wedding.}}\\

    \hline  
    \end{tabular}
    \caption{Analysis of the skeleton extraction module. Given a whole sentence as input, the words in red are the extracted skeleton.  }
    \label{tab:ex-sk}
\end{table}

\subsection{Error Analysis}

Although the proposed model outperforms the state-of-the-art models, it needs to be noted that the highest coherence score, 5.62, is a moderate result in human evaluation, indicating that there is still a long way to go before the generated stories reach the human level. Therefore, in this subsection, we give a detailed error analysis to explore what factors affect the performance.

\begin{figure}[t!]

\footnotesize
\centering
\def\angle{0}
\def\radius{3}
\def\cyclelist{{"red","green","blue","yellow"}}
\newcount\cyclecount \cyclecount=-1
\newcount\ind \ind=-1
\begin{tikzpicture}[nodes = {font=\sffamily \small},scale=0.45]
  \foreach \percent/\name in {
      62.8/Irrelevant Scenes,
      25.8/Chaotic Syntax,
      5.7/Chaotic timeline,
      5.7/Repeated Scenes,
    } {
      \ifx\percent\empty\else               
        \global\advance\cyclecount by 1     
        \global\advance\ind by 1            
        \ifnum5<\cyclecount                 
          \global\cyclecount=0              
          \global\ind=0                     
        \fi
        \pgfmathparse{\cyclelist[\the\ind]} 
        \edef\color{\pgfmathresult}         
        \draw[fill={\color!70!white},draw={\color}] (0,0) -- (\angle:\radius)
          arc (\angle:\angle+\percent*3.6:\radius) -- cycle;
        \node at (\angle+0.5*\percent*3.6:0.7*\radius) {\percent\,\%};
        \node[pin={[pin distance=5pt]\angle+0.5*\percent*3.6:\name}]
          at (\angle+0.5*\percent*3.6:\radius) {};
        \pgfmathparse{\angle+\percent*3.6}  
        \xdef\angle{\pgfmathresult}         
      \fi
    };
\end{tikzpicture}
\vspace{-2.0\baselineskip}
\caption{The distribution of error types.}
\label{fig: error type}

\end{figure}
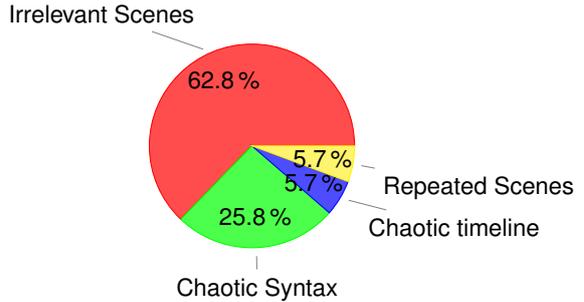

First, we classify the generated stories with scores below 6 that are considered less coherent. We conclude 4 types of errors from these outputs and the distribution of error types are shown in Figure~\ref{fig: error type}. It is expected that the irrelevant scenes make up most of the errors. In addition, there are several examples that are hard to be understood due to chaotic syntax.  For the type of chaotic timeline, the model neglects the time order of scenes and the generated stories goes backward in time. The repeated scenes mean that the generated stories just describe the input again. The above errors show that there are many dimensions in coherence, including scene-specific relevance, temporal connection, and non-recurrence. Modeling such dimensions is still a hard problem.

\begin{figure}[t]
\centering
\begin{minipage}[t]{0.49\linewidth}
\centering
\includegraphics[width=4.2cm]{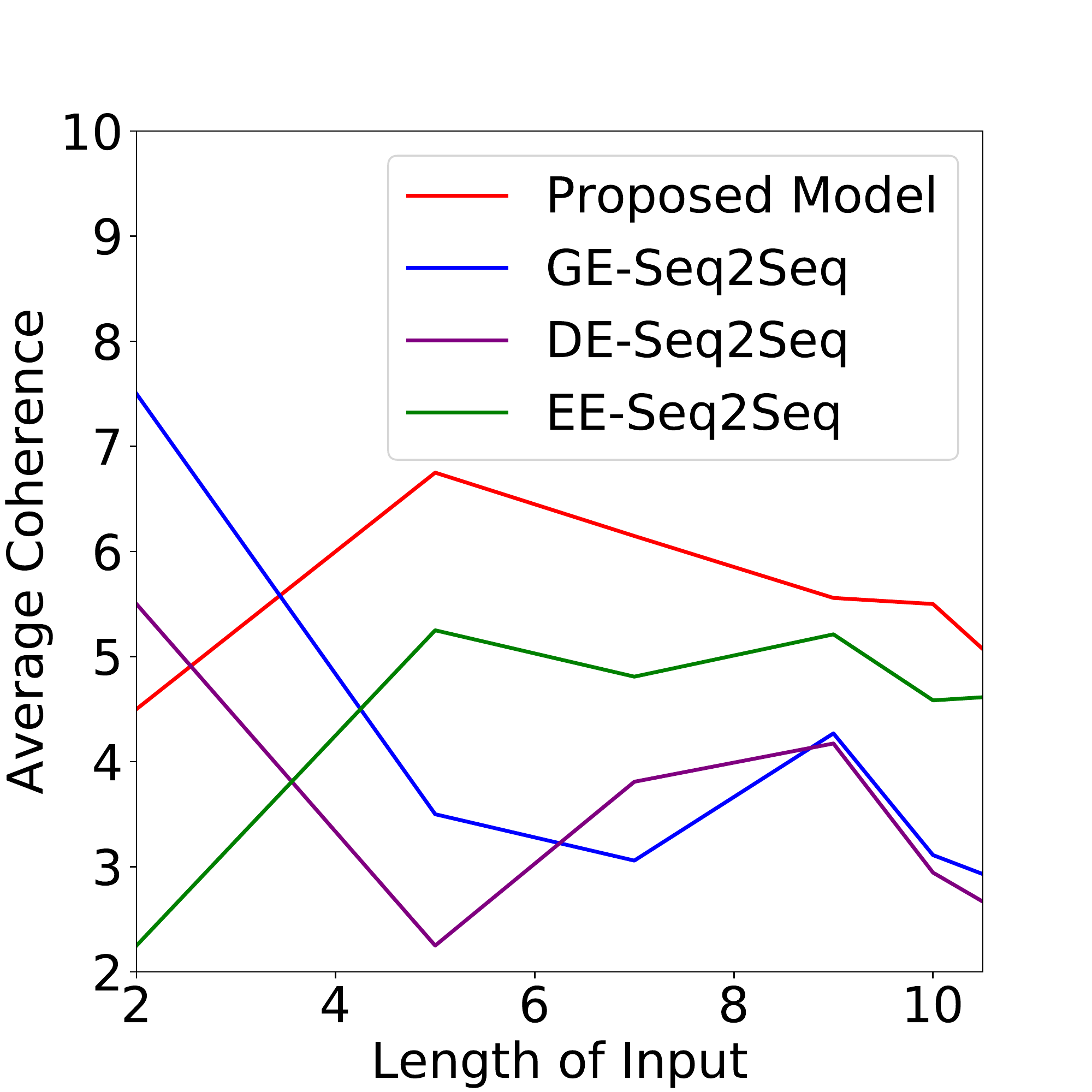}
\end{minipage}
\begin{minipage}[t]{0.49\linewidth}
\centering
\includegraphics[width=4.2cm]{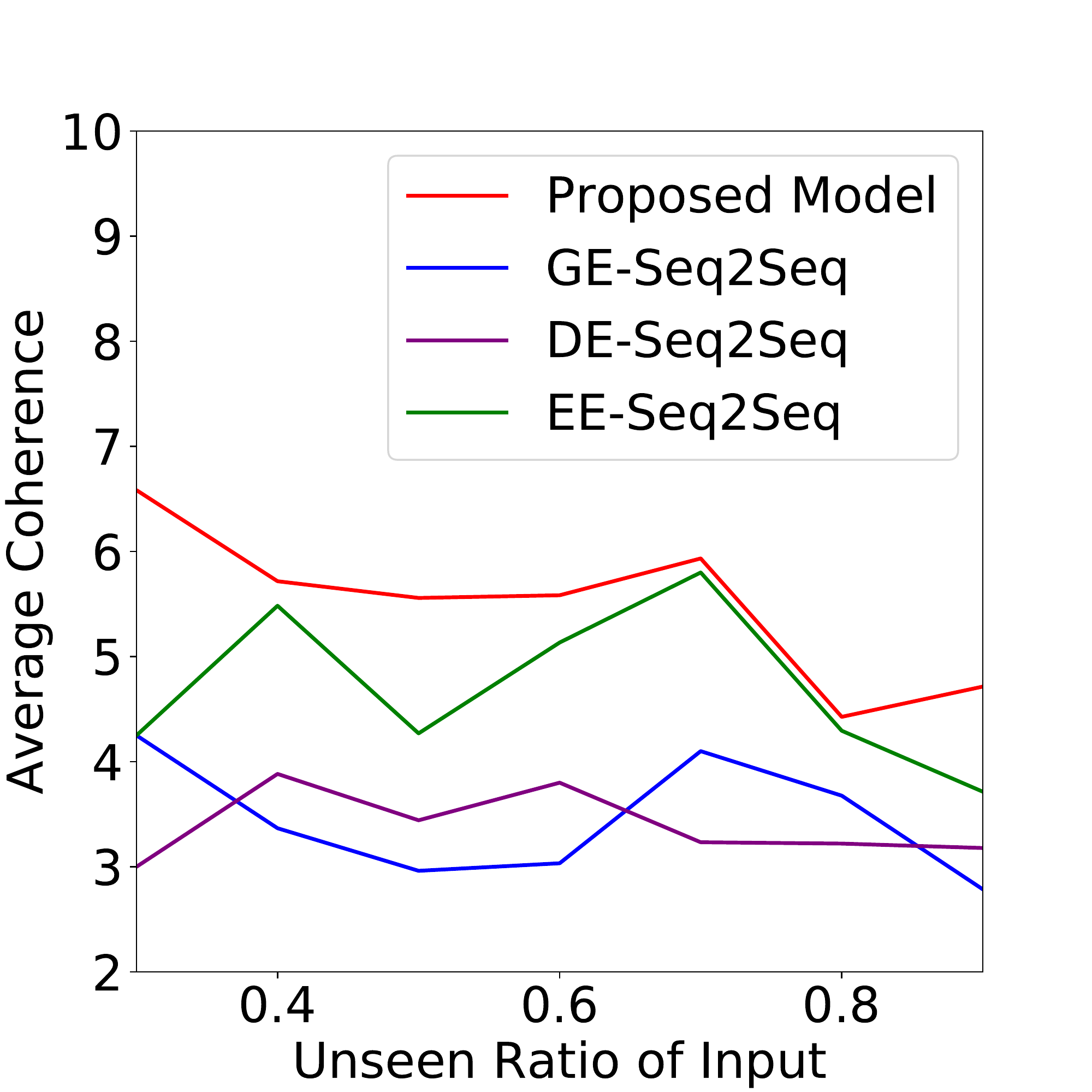}
\end{minipage}

\caption{ An illustration of how the performance is affected by the length of input (left) and the unseen ratio of input (right).  }

\vspace{-1.0\baselineskip}
\label{fig:error} 
\end{figure}

Furthermore, we explore how the performance is affected by the length of input and the unseen ratio of input. The results are shown in Figure~\ref{fig:error}. ``Unseen ratio'' is the percentage of the phrases that are not seen in the training data. We use the gap between $1$ and the BLEU score with the training data as the reference to compute it. When the input is short and the model often sees the input, the generated story tends to have high coherence. However, when the length of input increases and the model is not familiar with the input, the coherence goes down. Since our model extracts the key semantics better, the dependency of sentences can be easier to learned, which brings the smaller decrease in coherence.




\section{Conclusion and Future Work}

In this work, we propose a new skeleton-based model for generating coherent narrative stories. Different from traditional models, the proposed model first generates a skeleton that contains the key information of a sentence, and then expands the skeleton to a complete sentence. Experimental results show that our model significantly improves the quality of generated stories, especially in coherence. However, even with the best human evaluation results, the error analysis shows that there are still many challenges in narrative story generation, which we would like to explore in the future.

\section*{Acknowledgements}

This work was supported in part by National Natural Science Foundation of China (No. 61673028). We thank all the reviewers for providing the constructive suggestions. Xu Sun is the corresponding author of this paper.

\bibliography{emnlp2018}
\bibliographystyle{acl_natbib_nourl}

\end{CJK}
\end{document}